\documentclass{article}
\usepackage{amsmath}

\usepackage[final]{neurips_2019}

\usepackage[utf8]{inputenc}
\usepackage[T1]{fontenc}
\usepackage{hyperref}
\usepackage{url}
\usepackage{booktabs}
\usepackage{amsfonts}
\usepackage{nicefrac}
\usepackage{microtype}
\usepackage{graphicx}
\usepackage{xcolor}
\usepackage{multirow}
\usepackage{subcaption}
\usepackage{enumitem}

\makeatletter
\renewcommand\section{\@startsection{section}{1}{\z@}%
  {-6pt \@plus -2pt \@minus -2pt}{1pt \@plus 1pt}%
  {\normalfont\large\bf\raggedright}}
\renewcommand\subsection{\@startsection{subsection}{2}{\z@}%
  {-4pt \@plus -2pt \@minus -2pt}{0.5pt \@plus 1pt}%
  {\normalfont\normalsize\bf\raggedright}}
\renewcommand\subsubsection{\@startsection{subsubsection}{3}{\z@}%
  {-3pt \@plus -1pt \@minus -1pt}{0.5pt \@plus 0.5pt}%
  {\normalfont\normalsize\bf\raggedright}}
\makeatother
\setlist{nosep,leftmargin=*}
\selectfont

\title{An Empirical Study of SFT--DPO Interaction and Parameterization in Small Language Models}

\author{
  Yuming Feng\thanks{These authors contributed equally.} \\
  Department of Computer Science \\
  Stanford University \\
  \texttt{yumingf@stanford.edu} \\
  \And
  Christy Yang\footnotemark[1] \\
  Department of Computer Science \\
  Stanford University \\
  \texttt{yangyx@stanford.edu} \\
}

\begin{document}

\maketitle

\begin{abstract}
Direct Preference Optimization (DPO) is widely used after supervised fine-tuning (SFT) to align language models, yet empirical behavior under small backbones and modest data is under-specified. We systematically compare SFT-only, DPO-only, and staged SFT$\rightarrow$DPO training alongside full fine-tuning (FFT) versus LoRA on a GPT-2--scale decoder, evaluating paraphrase detection and Shakespearean sonnet continuation. DPO yields small, task-dependent gains over strong SFT and can match competitive SFT accuracy without a warm start when the preference construction closely parallels the supervised objective. In contrast, parameterization dominates: FFT consistently outperforms LoRA at matched training depth, and LoRA does not reduce wall-clock time on our hardware. These findings indicate that, in this small-scale regime, supervised full-parameter adaptation remains the primary performance lever, while preference optimization and low-rank adaptation provide limited marginal returns.
\end{abstract}

\vspace{-0.3em}
\begin{center}
\small\textit{Code and experiments:}~\href{https://github.com/Harry20030331/cs224n_project}{\texttt{github.com/Harry20030331/cs224n\_project}}
\end{center}
\vspace{0.3em}

\section{Introduction}
Pretrained language models can be adapted to downstream tasks through fine-tuning, but efficient adaptation remains challenging for smaller models with limited compute and parameter budgets. Two widely used approaches address this: \textit{parameter-efficient fine-tuning} such as Low-Rank Adaptation (LoRA), which updates few parameters while freezing most weights, and \textit{preference-based optimization} such as Direct Preference Optimization (DPO)~\cite{rafailov2024directpreferenceoptimizationlanguage}, which aligns models with desired outputs. In practice, models are often first trained with supervised fine-tuning (SFT) and then refined with DPO.

Despite widespread use, these techniques are incompletely understood in the small-model regime. Two questions remain open: (1) how LoRA~\cite{hu2021lora}  compares to full fine-tuning (FFT) when adapting smaller models, and (2) how SFT and DPO interact and when to introduce DPO during training. In this work we investigate both. We compare FFT with LoRA-based adaptation and examine the SFT--DPO interaction within a staged pipeline, analyzing how preference optimization behaves when applied after different SFT stages. We evaluate on paraphrase detection and sonnet generation using a GPT-2 backbone, providing insight into practical strategies for adapting smaller language models.

\vspace{3mm}
Our main contributions are:
\begin{itemize}
    \item We present a controlled empirical study of SFT-only, DPO-only, and staged SFT$\rightarrow$DPO training on a GPT-2--scale decoder, including DPO hyperparameter sweeps and handoff timing on paraphrase detection, plus preference-pair designs for sonnet continuation.
    \item We benchmark FFT against LoRA (rank sweep, training curves, and wall-clock) under matched task and data conditions, showing that parameterization differences dominate the gains from adding a preference stage in our regime.
    \item We distill practical implications for small models and modest data: full-parameter SFT remains the primary accuracy lever, DPO yields only marginal improvements over strong SFT, and LoRA does not translate into faster training on compute-bound hardware at this scale.
\end{itemize}

\section{Related Work}

\paragraph{Language Model Fine-Tuning}

Large pretrained language models have demonstrated strong performance across a wide range of tasks when adapted through fine-tuning. Early work such as GPT-2 \cite{radford2019language} showed that large autoregressive language models trained on large corpora can perform many downstream tasks with minimal task-specific modifications. Subsequent research has further explored how pretrained models can be adapted efficiently to new tasks through SFT and instruction tuning \cite{chung2022scalinginstructionfinetunedlanguagemodels}. These approaches form the foundation of most modern LLM training pipelines.

\paragraph{Preference-Based Optimization}

Beyond supervised learning, recent work has explored aligning language models using human preference signals. Reinforcement learning from human feedback (RLHF) has become a widely adopted framework for aligning LLMs with human intent \cite{ouyang2022training}. Earlier work also investigated learning from human feedback in sequence generation settings \cite{kreutzer2018reliabilitylearnabilityhumanbandit}. More recently, DPO \cite{rafailov2024directpreferenceoptimizationlanguage} proposes a simpler alternative to RLHF that directly optimizes preference pairs without explicitly training a reward model. DPO has become an increasingly popular alignment method due to its conceptual simplicity and empirical effectiveness.

\paragraph{Parameter-Efficient Fine-Tuning}

Another line of research focuses on reducing the computational cost of fine-tuning large models. Parameter-efficient fine-tuning methods aim to update only a small subset of parameters while keeping most pretrained weights fixed. Among these approaches, LoRA introduces low-rank adaptation matrices that allow efficient task adaptation with a small number of trainable parameters. Such methods are particularly attractive for large-scale models where FFT is computationally expensive.

\paragraph{Our Work}

While preference optimization and parameter-efficient fine-tuning are both widely used in modern language model training pipelines, their behavior in smaller model settings remains less explored. In particular, it is unclear how LoRA compares to FFT when adapting smaller language models, and how SFT interacts with preference optimization methods such as DPO. In this work, we study these questions using a GPT-2 backbone across both classification and generation tasks, focusing on the interaction between SFT and DPO as well as the practical impact of parameterization choices.

\section{Task Formulation}

We evaluate using two tasks that cover classification and generation settings.

\paragraph{Paraphrase Detection}
Our classification task is paraphrase detection on the Quora Question Pairs dataset. Given a pair of questions, the model predicts whether they express the same meaning. This setting provides a large labeled dataset and requires semantic matching beyond surface overlap, and we use it to study parameterization strategies (FFT vs.\ LoRA) and the interaction between SFT and DPO.

\paragraph{Sonnet Generation}
To complement the classification experiments, we include a generation task based on Shakespearean sonnet continuation. Given the prefix of a sonnet, the model autoregressively generates the remaining lines. This task enables evaluation of open-ended generation behavior and provides a natural setting for constructing preference pairs used in preference-based optimization.

\section{Approach}
\label{sec:approach}

In this section, we describe the model architecture and training methods used in our study. Our goal is to analyze how different training objectives and parameterization strategies influence the adaptation of a pretrained language model. In particular, we investigate the interaction between SFT and DPO, as well as the comparison between FFT and LoRA.

\subsection{Base Model}

Our backbone model is GPT-2 (124M parameters), a decoder-only Transformer language model~\cite{radford2019language}. The model consists of token embeddings and positional embeddings followed by a stack of Transformer blocks with causal self-attention and feed-forward networks.

Each Transformer block computes
\begin{equation}
\hat{h} = h + \mathrm{Attn}(\mathrm{LN}(h)), \qquad
h' = \hat{h} + \mathrm{FFN}(\mathrm{LN}(\hat{h}))
\end{equation}
where $\mathrm{Attn}$ denotes multi-head causal self-attention and $\mathrm{FFN}$ denotes the feed-forward network with GELU activation. The hidden size is $d=768$ with 12 attention heads and 12 layers. 

We implement the GPT-2 architecture and load pretrained weights from HuggingFace as initialization.

\subsection{Task Adaptation}

The pretrained GPT-2 model is adapted differently depending on the task type.

For paraphrase detection, we attach a linear head on the final-token hidden state to produce label logits. For generation (sonnet continuation), the model generates autoregressively via the language modeling head (hidden states projected onto the token embedding matrix).

\subsection{Supervised Fine-Tuning}

We first adapt the pretrained model using SFT. Given an input $x$ and target output $y$, the model parameters $\theta$ are optimized using the standard cross-entropy loss

\begin{equation}
\mathcal{L}_{\text{SFT}} = - \log P_{\theta}(y \mid x).
\end{equation}

SFT serves as the primary baseline training objective and provides the initial model for subsequent preference-based optimization.

\subsection{Direct Preference Optimization}

To incorporate preference-based learning, we explore DPO~\cite{rafailov2024directpreferenceoptimizationlanguage}. DPO trains the model to prefer a chosen response over a rejected response given the same prompt.

Given a prompt $x$, a preferred response $y_w$, and a rejected response $y_l$, the objective encourages the model to assign higher probability to the preferred response. The loss is defined as

\begin{equation}
\mathcal{L}_{\text{DPO}} =
- \log \sigma \!\left(
\beta \left(
\log P_{\theta}(y_w \mid x) - \log P_{\theta}(y_l \mid x)
\right)
\right),
\end{equation}

where $\sigma$ is the sigmoid function and $\beta$ controls the strength of the preference signal.

Preference pairs use correct vs.\ incorrect labels for paraphrase detection, and reference vs.\ model-generated continuations for generation. We compare SFT-only, DPO-only, and a two-stage SFT$\rightarrow$DPO pipeline.

\subsection{Parameterization Strategies}

To study the effect of parameterization strategies, we compare two approaches for adapting the pretrained model: FFT and LoRA.

\subsubsection{Full Fine-Tuning}

In FFT, all model parameters are updated during training. This approach provides maximum flexibility but requires updating the entire parameter set of the model.

\subsubsection{LoRA}

LoRA~\cite{hu2021lora} introduces trainable low-rank matrices to approximate weight updates while keeping the pretrained weights frozen. Specifically, for a weight matrix $W$, the update is parameterized as

\begin{equation}
W = W_0 + \frac{\alpha}{r} BA
\end{equation}

where $A \in \mathbb{R}^{r \times d}$ and $B \in \mathbb{R}^{d \times r}$ are low-rank matrices and $r$ is the rank of the adaptation. This significantly reduces the number of trainable parameters while maintaining competitive performance.

\section{Experiments}

\subsection{Data}
This section describes the datasets used for our tasks and the procedure for constructing preference pairs used in DPO training.
\subsubsection{Task Datasets}

For paraphrase detection, we use the Quora Question Pairs dataset~\cite{quora2017questionpairs} with 283,011 train, 40,430 dev, and 80,860 test examples.

For sonnet generation, we use the Folger Shakespeare Library edition of Shakespeare's 155 sonnets. The dataset is split into 131 training poems, 12 development poems, and 12 test poems. Each 14-line sonnet is divided into a 3-line conditioning prompt and an 11-line target continuation.

\subsubsection{Preference Pair Construction}

For DPO training, we construct preference pairs consisting of a prompt, a preferred response, and a rejected response.

For paraphrase detection, preference pairs are derived directly from labeled data. Given an input example, the correct label is treated as the preferred output while the incorrect label is treated as the rejected output. This allows DPO to increase the log-probability margin between the correct and incorrect classes.

For the sonnet generation task, the preferred response is the original Shakespeare continuation and the rejected response is generated by the model. The prompt consists of the first three lines of a sonnet, and the model generates candidate continuations for the remaining lines.

We explore three strategies for constructing preference pairs:

\paragraph{V1: Full-overlap pairs.}  
Preference pairs are generated from the same 131 sonnets used for SFT. For each prompt, we sample 10 candidate continuations from the model. Candidates are filtered using a chrF similarity band with respect to the reference poem. Specifically, we keep candidates whose chrF scores fall within the range $[60, 90]$ to remove degenerate outputs and near-identical generations. Among the remaining candidates, the lowest-scoring continuation is selected as the rejected response. This procedure yields 97 preference pairs.

\paragraph{V2: Data-split pairs.}  
To avoid generating continuations from prompts seen during supervised training, we split the dataset into two halves. One half is used for SFT training, while preference pairs are constructed from prompts in the other half. The same sampling and filtering procedure is applied. This produces 65 preference pairs from unseen prompts.

\paragraph{V3: Top-$K$ augmented pairs.}  
To increase the number of preference pairs, we further augment the dataset by selecting multiple rejected candidates for each prompt. Using the same train/dev split as in V2, we keep the top five filtered candidates per prompt to construct additional pairs. This results in 325 preference pairs from 65 unique prompts.

\subsection{Evaluation method}
Paraphrase detection is evaluated using dev-set \textbf{accuracy} (primary metric for model selection) along with macro-averaged precision, recall, and F1.
Sonnet generation is evaluated using \textbf{chrF}~\cite{popovic2015chrf}, a character-level $n$-gram F-score computed by sacrebleu, which is more robust than BLEU on small corpora and captures partial character-level overlap that word-level metrics miss. Because sampling-based generation is inherently stochastic, we additionally run \textbf{stability evaluations}: each checkpoint is evaluated with 20 different random seeds at the chosen temperature, and we report the mean, standard deviation, minimum, and maximum chrF across those 20 runs, so that reported gains can be compared against seed-to-seed variance.

\subsection{Experimental Details}

All experiments were run on NVIDIA H100 GPUs using the Modal cloud platform (80GB VRAM per GPU).

Training time differs significantly between tasks. For the paraphrase detection task, training a single epoch typically takes approximately 7 minutes. In contrast, the sonnet generation experiments are much smaller in scale and require only about 0.7 seconds per training epoch.

For the main training pipeline (SFT and DPO), we extend a PyTorch GPT-2 fine-tuning stack with DPO and staged SFT$\rightarrow$DPO handoffs.\footnote{Public repository: \url{https://github.com/Harry20030331/cs224n_project}.}

For parameter-efficient fine-tuning experiments, we implement LoRA using the Hugging Face ecosystem together with the PEFT (Parameter-Efficient Fine-Tuning) library.

Detailed hyperparameter settings and task-specific configurations for each experiment are provided in the Appendix.

\subsection{Results on Paraphrase Detection}
\label{sec:paraphrase-results}

We evaluate the interaction between training objectives and parameterization strategies on the Quora Question Pairs paraphrase detection task. Unless otherwise specified, all results reported in this section are best development-set results, and model selection is performed without using test labels.

\subsubsection{Effect of Data Size}
\label{sec:data-size}

We compare three dataset sizes (2.83k, 28.3k, 283k) to assess the effect of training data scale (Table~\ref{tab:data-size}).
\vspace{-0.6em}
\begin{table}[h]
\centering
\footnotesize
\caption{Effect of training data size on paraphrase detection. Larger datasets improve development performance and reduce the train--dev loss gap.}
\begin{tabular}{lccccccccc}
\toprule
Data size & Epochs & Time/epoch & Total time & Train loss & Dev loss & Acc. & Prec. & Rec. & F1 \\
\midrule
2.83k  & 40 & 30.25s & 20.2 min & 0.033 & 0.770 & 78.7 & 77.1 & 77.3 & 77.2 \\
28.3k  & 15 & 86.16s & 21.5 min & 0.123 & 0.403 & 83.9 & 82.9 & 82.3 & 82.6 \\
283k   & 8  & 435s   & 58.0 min & 0.200 & 0.280 & 88.7 & 87.5 & 89.3 & 88.2 \\
\bottomrule
\end{tabular}
\label{tab:data-size}
\end{table}

\vspace{-0.6em}
\begin{table}[h]
\centering
\footnotesize
\caption{Same-time training comparison between large and small datasets. Under a fixed training-time budget, fewer epochs on a larger dataset outperform many epochs on a smaller dataset.}
\begin{tabular}{lcccc}
\toprule
Setting & Training Time & Configuration & Dev F1 & Dev Accuracy \\
\midrule
\multirow{2}{*}{$\sim$15 min} 
& $\sim$14.5 min & 283k $\times$ 2 epochs & 86.20 & 86.81 \\
& $\sim$14 min & 28.3k $\times$ 10 epochs & 82.34 & 83.25 \\
\midrule
\multirow{2}{*}{$\sim$30 min} 
& $\sim$29 min & 283k $\times$ 4 epochs & 87.11 & 87.94 \\
& $\sim$29 min & 28.3k $\times$ 20 epochs & $\sim$82.9 & $\sim$83.9 \\
\bottomrule
\end{tabular}
\label{tab:same-time}
\end{table}

Larger datasets consistently improve dev performance (Table~\ref{tab:data-size}). At 2.83k the model overfits severely (train loss 0.033 vs.\ dev loss 0.770); at 283k dev F1 reaches 88.2. Under a fixed training-time budget (Table~\ref{tab:same-time}), fewer epochs on a larger dataset outperform more epochs on a smaller one---e.g., 283k$\times$2 epochs in $\sim$15 min yields 86.20 F1 vs.\ 28.3k$\times$10 epochs at 82.34 F1. Data diversity is more valuable than repeated exposure: the model learns the task structure quickly, so more distinct examples generalize better. We adopt the full 283k dataset for all subsequent experiments.

\subsubsection{Effect of Parameterization}
\label{sec:param-results}

We compare FFT against LoRA on the full 283k dataset for 3 epochs, sweeping LoRA rank $r \in \{4, 8, 16\}$ (Table~\ref{tab:lora-rank}).
\vspace{-1.2em}
\begin{table}[h]
\centering
\footnotesize
\caption{Comparison of FFT and LoRA on paraphrase detection at epoch 3.}
\begin{tabular}{lcccccc}
\toprule
Method & Config & Epochs & Acc. & Prec. & Rec. & F1 \\
\midrule
FFT (SFT) & full  & 3 & 89.24 & 88.34 & 88.65 & 88.49 \\
LoRA      & $r=16$ & 3 & 85.15 & 83.99 & 85.84 & 84.54 \\
LoRA      & $r=8$  & 3 & 86.34 & 85.12 & 86.18 & 85.56 \\
LoRA      & $r=4$  & 3 & 85.67 & 84.54 & 86.48 & 85.09 \\
\bottomrule
\end{tabular}
\label{tab:lora-rank}
\end{table}

FFT outperforms all LoRA settings. Among LoRA variants, $r=8$ performs best while $r=16$ performs worse than $r=8$ and $r=4$---a rank paradox we attribute to optimization: at epoch 3, the higher-rank adapter introduces more parameters that receive the same number of updates and are not effectively utilized. We adopt $r=8$ as the default LoRA configuration.

\begin{figure}[h]
    \centering
    \includegraphics[width=0.92\linewidth]{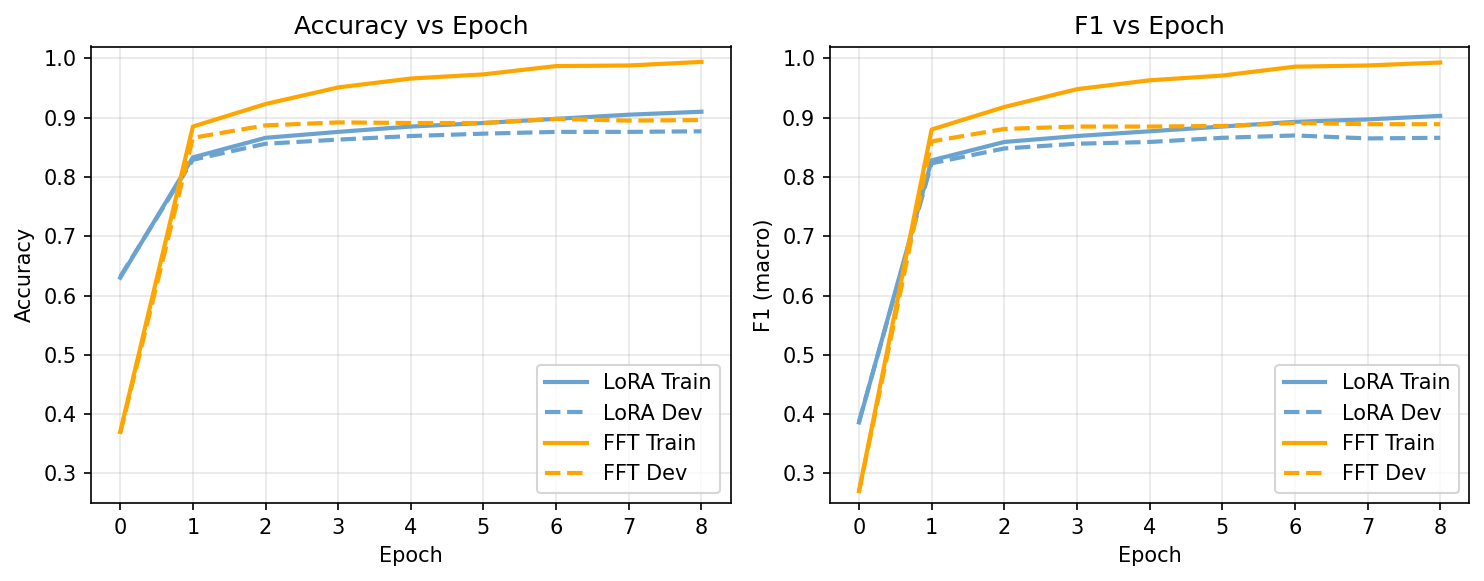}
    \vspace{-4pt}
    \caption{Training curves for FFT and LoRA ($r=8$) on paraphrase detection. Plot train and dev accuracy/F1 across epochs.}
    \label{fig:fft-lora-curves}
\end{figure}

To better understand the optimization dynamics of FFT and LoRA, we also ran an extended 8-epoch comparison using FFT and LoRA with $r=8$. We summarize these trajectories in Figure~\ref{fig:fft-lora-curves}. The figure should plot train/dev accuracy and F1 across epochs for both methods.

Both methods improve sharply in the first epoch then slow down. FFT overfits more (train keeps improving after dev plateaus); LoRA has a smaller train--dev gap but a lower dev ceiling.

Another practical motivation for using LoRA is training efficiency. In principle, LoRA reduces the number of trainable parameters and memory footprint, which can allow larger batch sizes and potentially faster training. However, in our experiments we did not observe a noticeable reduction in training time when using LoRA compared with FFT.

Our H100 setup is compute-bound rather than memory-bound, so LoRA's reduced memory footprint does not translate to faster training. For larger models, LoRA may offer clearer efficiency gains; here, its main benefit is memory savings.



\subsubsection{Effect of DPO Hyperparameters}
\label{sec:dpo-hparams}

Before studying the SFT$\rightarrow$DPO transition, we first tune the DPO hyperparameters. 
All runs in this subsection start from the best FFT SFT checkpoint (SFT@9, i.e., the checkpoint after the ninth SFT epoch), use the full 283k dataset, and construct preference pairs as chosen = correct label and rejected = incorrect label.

We first sweep the DPO learning rate while fixing $\beta=0.2$, and then sweep $\beta$ while fixing the best learning rate. Results are shown in Table~\ref{tab:dpo-hparams}.The learning rate choice is not highly sensitive in Table~\ref{tab:dpo-hparams}, but the smallest value performs slightly better. This is consistent with DPO acting as a refinement stage on top of a strong SFT initialization: a smaller learning rate adjusts margins without overshooting. Similarly, the $\beta$ sweep shows that $\beta=0.2$ performs best, although the differences are modest. A smaller $\beta$ weakens the preference signal, while a larger $\beta$ amplifies the margin between chosen and rejected responses, which can lead to overly aggressive updates and slightly worse dev performance. We therefore use learning rate $5\times10^{-6}$ and $\beta=0.2$ in all subsequent DPO experiments.
\vspace{-0.6em}
\begin{table}[h]
\centering
\footnotesize
\caption{DPO hyperparameter sweeps on paraphrase detection.}

\begin{subtable}{0.48\linewidth}
\centering
\begin{tabular}{lcccc}
\toprule
LR & Acc & Prec & Rec & F1 \\
\midrule
$5\times10^{-6}$ & 89.75 & 88.71 & 89.67 & 89.13 \\
$1\times10^{-5}$ & 89.73 & 88.66 & 89.77 & 89.13 \\
$2\times10^{-5}$ & 89.61 & 88.57 & 89.49 & 88.97 \\
\bottomrule
\end{tabular}
\caption{Learning rate sweep ($\beta=0.2$).}
\end{subtable}
\hfill
\begin{subtable}{0.48\linewidth}
\centering
\begin{tabular}{lcccc}
\toprule
$\beta$ & Acc & Prec & Rec & F1 \\
\midrule
0.1 & 89.72 & 88.67 & 89.67 & 89.10 \\
0.2 & 89.75 & 88.71 & 89.67 & 89.13 \\
0.5 & 89.69 & 88.64 & 89.61 & 89.06 \\
\bottomrule
\end{tabular}
\caption{$\beta$ sweep (LR $=5\times10^{-6}$).}
\end{subtable}

\label{tab:dpo-hparams}
\end{table}

\subsubsection{Effect of SFT$\rightarrow$DPO Handoff}
\label{sec:handoff}

We next study when DPO should be introduced relative to supervised training. 
We compare four strategies: SFT only, DPO only, SFT@3$\rightarrow$DPO, and SFT@9$\rightarrow$DPO. 
Here SFT@N denotes the checkpoint after N epochs of SFT. 
We choose SFT@3 as an early-stage checkpoint where training progress begins to slow, and SFT@9 as a near-converged checkpoint. 
In the mixed strategies, DPO starts from the corresponding SFT checkpoint. 
Results are shown in Table~\ref{tab:handoff}.

\begin{table}[h]
\centering
\footnotesize
\caption{Effect of SFT$\rightarrow$DPO handoff timing on paraphrase detection.}
\begin{tabular}{lccccccc}
\toprule
Strategy & SFT ep & DPO ep & Best & Acc & Prec & Rec & F1 \\
\midrule
SFT only & 10 & 0 & SFT@9 & 89.87 & 88.89 & 89.59 & 89.21 \\
DPO only & 0 & 10 & DPO@7 & 89.46 & 88.38 & 89.43 & 88.83 \\
SFT@3$\rightarrow$DPO & 3 & 10 & DPO@9 & 89.77 & 88.78 & 89.54 & 89.12 \\
SFT@9$\rightarrow$DPO & 9 & 6 & DPO@5 & 90.05 & 89.05 & 89.91 & 89.43 \\
\bottomrule
\end{tabular}
\label{tab:handoff}
\end{table}

The strongest result in Table~\ref{tab:handoff} is obtained by first training SFT to the best checkpoint and then running DPO: SFT@9$\rightarrow$DPO reaches 90.05\% accuracy and 89.43 F1.

At the same time, the differences across strategies are small. The spread in best dev accuracy is less than 0.6 points (89.46–90.05). Since each setting is run only once, we treat these differences as descriptive rather than statistically significant.

Interestingly, DPO-only is competitive with SFT-only. This differs from the usual intuition that DPO requires a strong SFT warm start. The likely reason is task structure: paraphrase detection is a fixed-prompt binary classification task, and our DPO objective uses the correct label as chosen and the wrong label as rejected. In this setting the preference signal closely resembles the supervised signal, allowing DPO to learn the decision boundary directly from scratch.

\subsubsection{Final Comparison}
\label{sec:final-2x2}

Finally, we summarize the best dev performance for the four cells of our study: training objective (SFT vs.\ DPO) $\times$ parameterization strategy (FFT vs.\ LoRA). For FFT we use the best handoff and DPO hyperparameters identified above. For LoRA we use rank $r=8$ and run DPO from the converged LoRA SFT checkpoint.

\begin{table}[h]
\centering
\footnotesize
\caption{Best dev results for training objective $\times$ parameterization on paraphrase detection.}
\begin{tabular}{lcc}
\toprule
Acc / F1 & SFT & DPO \\
\midrule
FFT  & 89.87 / 89.21 & 90.05 / 89.43 \\
LoRA & 87.70 / 87.00 & 88.48 / 87.76 \\
\bottomrule
\end{tabular}
\label{tab:final-results}
\end{table}

Three conclusions follow from Table~\ref{tab:final-results}. First, FFT consistently outperforms LoRA regardless of the training objective, confirming that full-model updates remain more effective than low-rank adaptation in this setting.

Second, DPO improves over SFT in both regimes. The gain is small for FFT (90.05 vs.\ 89.87 accuracy) but larger for LoRA (88.48 vs.\ 87.70), suggesting that DPO provides a mild benefit by increasing the separation between correct and incorrect outputs.

Third, the effect of parameterization is larger than the effect of objective. Moving from LoRA to FFT changes accuracy by more than one point, whereas moving from SFT to DPO changes accuracy by less than one point. In practical terms, the choice between FFT and LoRA matters more than the choice between SFT and DPO for this task.

\subsection{Results on Sonnet Generation}
\label{sec:sonnet-results}

To complement the classification experiments, we also study a generative task: Shakespearean sonnet continuation. Here the goal is open-ended continuation quality rather than binary decisions, measured using chrF on the dev set. We analyze two factors: sampling temperature and DPO preference-pair construction.

\subsubsection{Effect of Sampling Temperature}
\label{sec:temperature}

We evaluate the best SFT checkpoint at several temperatures (20 seeds each) to test whether lower temperature improves chrF by making generation more deterministic.

\vspace{-0.6em}
\begin{table}[h]
\centering
\footnotesize
\caption{Effect of sampling temperature on sonnet-generation chrF across 20 seeds.}
\begin{tabular}{lcccc}
\toprule
Temp & Mean & Std & Min & Max \\
\midrule
0.5 & 41.29 & 0.255 & 40.94 & 42.10 \\
1.0 & 41.54 & 0.362 & 40.44 & 42.05 \\
1.2 & 41.70 & 0.308 & 41.20 & 42.32 \\
1.5 & 41.86 & 0.280 & 41.20 & 42.42 \\
2.0 & 41.78 & 0.573 & 40.09 & 42.45 \\
\bottomrule
\end{tabular}
\label{tab:temperature}
\end{table}

The results contradict the simple ``lower temperature is better'' hypothesis. The lowest temperature ($T=0.5$) produces the worst mean chrF. Performance improves as temperature increases, peaks around $T=1.5$, and becomes less stable at $T=2.0$.

Low temperatures make generation overly deterministic, often producing repetitive or conservative continuations that reduce overlap with the Shakespeare reference. Increasing the temperature allows more lexical diversity while still preserving coherence, which improves average chrF. Although higher temperatures typically increase sampling variance, in our setup decoding also uses nucleus sampling with $p=0.9$, which truncates the low-probability tail of the distribution. As a result, even $T=1.5$ does not produce extremely unlikely tokens, allowing moderate temperature values to improve diversity without severely degrading quality. We therefore report sonnet results using temperatures 1.5.

\subsubsection{Effect of Preference Pair Construction}
\label{sec:pair-construction-results}

We compare three strategies for constructing DPO preference pairs: V1 (same sonnets as SFT), V2 (50--50 split, DPO on unseen prompts), and V3 (top-$K$ augmentation, 325 pairs from 65 prompts). Results are in Table~\ref{tab:pair-construction}.

\begin{table}[h]
\centering
\footnotesize
\caption{Comparison of DPO preference-pair construction strategies for sonnet generation.}
\begin{tabular}{lcccc}
\toprule
Config & Pairs & Best chrF & Epoch & $\Delta$ vs SFT \\
\midrule
DPO V1 & 97 & 41.94 & 2 & +0.19 \\
DPO V2 & 65 & 41.74 & 4 & +0.28 \\
DPO V3 & 325 & 41.46 & 0 & 0 \\
\bottomrule
\end{tabular}
\label{tab:pair-construction}
\end{table}

DPO provides only minor gains over the SFT baseline on this task. V1 yields a small improvement, V2 is roughly comparable, and V3 collapses.

In V1 the preference pairs are constructed from the same 131 sonnets used for SFT, so DPO mainly reinforces a signal the model has already learned. V2 uses unseen prompts, which reduces memorization concerns but further reduces the already small dataset. V3 has the largest number of pairs (325) but only 65 unique prompts, meaning the model repeatedly sees the same prompts with slightly different rejected continuations. Rather than increasing useful diversity, this over-reinforces a narrow prompt set and leads to unstable training.

Taken together, these results suggest that preference optimization is limited in extremely low-resource settings. In our case both the model scale and the dataset size are small, leaving little room for DPO to meaningfully reshape the model’s behavior beyond what SFT already provides. More broadly, this points to a practical regime in which preference tuning becomes effective only when model capacity and data scale are sufficiently large.

\section{Analysis}
\subsection{Qualitative Error Analysis on Paraphrase Detection}
\label{sec:error_analysis}

\begin{table}[h]
  \centering
  \small
  \caption{Paraphrase detection Case Study}
  \label{tab:paraphrase-qualitative}
  \begin{tabular}{@{}lp{3.6cm}p{3.6cm}ccp{1.5cm}@{}}
    \toprule
    Case    & Sentence 1 & Sentence 2 & Gold & Model & Result \\
    \midrule
    Success & Is it biologically good or bad to marry other caste?
            & Is it good or bad to marry other caste?
            & Yes  & Yes  & Correct \\
    \addlinespace
    Failure & Why do Quorans answer questions that are already answered?
            & Why do Quorans downvote questions they cannot answer?
            & No   & Yes  & Incorrect (FP) \\
    \bottomrule
  \end{tabular}
\end{table}

The success example is a near-duplicate pair differing only by the word “biologically”; the model correctly predicts paraphrase. The failure is a false positive: both sentences share topic and wording (Quorans, questions) but ask different things (answering already-answered questions vs. downvoting questions one cannot answer). Taken together, these cases indicate that the small model’s capacity for fine-grained semantic understanding is still limited.
\subsection{Qualitative comparison: SFT continuation vs. original Sonnet 132}
\label{sec:sonnet_quali_comparison}

The best-checkpoint SFT–fine-tuned GPT-2 continuation for Sonnet 132 (full gold and generated texts in Appendix~\ref{app:sonnet132}) successfully imitates several surface-level stylistic properties of Shakespearean verse. It employs elevated lexis (“languished,” “gentle praise”), archaic or formal constructions, and thematically appropriate imagery centered on lips, speech, and affection, which broadly aligns with the emotional register of the original poem. This indicates that SFT effectively steers the model toward the target style at the level of local word choice and short-span phrasing.

In contrast, a direct comparison with the gold continuation reveals persistent weaknesses in global structure and formal control. The original sonnet develops a coherent conceit in which the beloved’s “mourning” eyes motivate a clear argument about beauty, pity, and complexion; the generated continuation, by comparison, drifts semantically and introduces syntactically awkward, partially uninterpretable lines (e.g., “Who liped in assent to kiss each other's burs”). Moreover, the model only loosely respects rhyme and meter, failing to consistently reproduce the iambic pentameter and end-rhyme pattern that characterize the Shakespearean sonnet form. Overall, this example suggests that while SFT enables GPT-2 to approximate local Shakespearean style, capturing long-range rhetorical organization and strict prosodic structure remains a significant challenge.

\section{Conclusion}
We isolate two design axes for adapting small decoder-only models: the SFT$\rightarrow$DPO training recipe and the choice between FFT and LoRA. Across paraphrase detection and sonnet generation, DPO is not a reliable large win over a well-tuned SFT baseline at this scale; gains are small and task-dependent, and DPO-from-scratch can be competitive when preferences align tightly with the supervised signal (e.g., chosen/rejected class labels). Parameterization matters more than the preference stage: FFT reliably achieves higher accuracy and chrF than LoRA, and LoRA does not translate into faster training under our compute-bound H100 setup.

We therefore conclude that, for GPT-2--class models and the data regimes we study, investing compute in full-parameter SFT and data scaling dominates marginal returns from DPO and low-rank adapters. Preference optimization and PEFT remain valuable tools at larger scales, but practitioners should not assume they replicate their ``large-model'' benefits when capacity and supervision are both scarce---explicit measurement on the target regime is warranted.

\clearpage
\bibliographystyle{unsrt}
\bibliography{references}

\appendix
\section*{Appendix}
\section{Hyperparameters}

This section summarizes the hyperparameters used in our experiments.

\begin{table}[!h]
\centering
\small
\caption{Hyperparameters for the dataset scale experiment.}
\begin{tabular}{lccc}
\toprule
Hyperparameter & 2.83k & 28.3k & 283k (chosen) \\
\midrule
Optimizer        & AdamW & AdamW & AdamW \\
Learning rate    & $1\times10^{-5}$ & $1\times10^{-5}$ & $1\times10^{-5}$ \\
Batch size       & 128   & 128   & 128 \\
Epochs           & 40    & 15    & 8 \\
Weight decay     & 0     & 0     & 0 \\
Max words/sent.  & 64    & 64    & 64 \\
\bottomrule
\end{tabular}
\end{table}

\begin{table}[!h]
\centering
\small
\caption{Hyperparameters for the FFT vs. LoRA rank experiment.}
\begin{tabular}{lcc}
\toprule
Hyperparameter & FFT (SFT) & LoRA \\
\midrule
Optimizer       & AdamW & AdamW \\
Learning rate   & $5\times10^{-5}$ & $2\times10^{-4}$ \\
Batch size      & 128 & 128 \\
Epochs          & 3 & 3 \\
Weight decay    & 0 & 0 \\
Max words/sent. & 64 & 64 \\
LoRA rank ($r$) & --- & 4, 8, 16 \\
LoRA $\alpha$   & --- & 16 \\
LoRA dropout    & --- & 0.0 \\
\bottomrule
\end{tabular}
\end{table}

\begin{table}[!h]
\centering
\small
\caption{DPO hyperparameters: search space (left) and hand-off phase (right).}
\label{tab:dpo-phase}
\begin{subtable}{0.48\linewidth}
\centering
\begin{tabular}{lc}
\toprule
Hyperparameter & Values \\
\midrule
Optimizer & AdamW \\
DPO learning rate & $5\!\times\!10^{-6}$, $1\!\times\!10^{-5}$, $2\!\times\!10^{-5}$ \\
Batch size & 128 \\
Weight decay & 0 \\
Max words/sent. & 64 \\
DPO $\beta$ & 0.1, 0.2, 0.5 \\
\bottomrule
\end{tabular}
\caption{Search space.}
\end{subtable}
\hfill
\begin{subtable}{0.48\linewidth}
\centering
\begin{tabular}{lc}
\toprule
Hyperparameter & Value \\
\midrule
Optimizer & AdamW \\
DPO learning rate & $5\times10^{-6}$ \\
Batch size & 128 \\
Weight decay & 0 \\
Max words/sent. & 64 \\
DPO $\beta$ & 0.2 \\
\bottomrule
\end{tabular}
\caption{Hand-off phase.}
\end{subtable}
\end{table}

\begin{table}[!h]
\centering
\small
\caption{Hyperparameters for the Sonnet Generation task (SFT and DPO phases).}
\begin{subtable}{0.48\linewidth}
\centering
\begin{tabular}{lc}
\toprule
Hyperparameter & Value \\
\midrule
Optimizer & AdamW \\
Epochs & 30 \\
Learning rate & $1\times10^{-5}$ \\
Batch size & 8 \\
Max tokens & 256 \\
Temperature & 1.5 \\
Top-$p$ & 0.9 \\
LoRA rank ($r$) & 8 \\
\bottomrule
\end{tabular}
\caption{SFT phase.}
\end{subtable}
\hfill
\begin{subtable}{0.48\linewidth}
\centering
\begin{tabular}{lc}
\toprule
Hyperparameter & Value \\
\midrule
Optimizer & AdamW \\
Epochs & 8 \\
DPO learning rate & $5\times10^{-6}$ \\
Batch size & 8 \\
DPO $\beta$ & 0.2 \\
Max tokens & 256 \\
Temperature & 1.5 \\
Top-$p$ & 0.9 \\
LoRA rank ($r$) & 8 \\
\bottomrule
\end{tabular}
\caption{DPO phase.}
\end{subtable}
\end{table}

\section{Case Study}
\subsection{Sonnet 132: Gold vs.\ SFT-Generated Continuation}
\label{app:sonnet132}

We provide the original held-out continuation of Sonnet 132 and the corresponding continuation generated by the best SFT checkpoint for qualitative comparison.

\begin{minipage}[t]{0.48\textwidth}
\raggedright
\textbf{Gold continuation (Sonnet 132).}

\vspace{0.5em}

Thine eyes I love, and they, as pitying me,\\
Knowing thy heart torment me with disdain,\\
Have put on black, and loving mourners be,\\
Looking with pretty ruth upon my pain.\\
And truly not the morning sun of heaven\\
Better becomes the gray cheeks of the east,\\
Nor that full star that ushers in the even\\
Doth half that glory to the sober west\\
As those two mourning eyes become thy face.\\
O, let it then as well beseem thy heart\\
To mourn for me, since mourning doth thee grace,\\
And suit thy pity like in every part.\\
Then will I swear beauty herself is black,\\
And all they foul that thy complexion lack.\\
\end{minipage}
\hfill
\begin{minipage}[t]{0.48\textwidth}
\raggedright
\textbf{SFT-generated continuation (best checkpoint).}

\vspace{0.5em}

Those lips that Love's own hand did make\\
Breathed forth the sound that said ``I hate''\\
To me that languished for her sake;\\
Who liped in assent to kiss each other's burs\\
As if by some instinct the two did know\\
Even how to kiss each other's cheeks; Who, no love, did as she talked,\\
Till they, tongue still moving, moved with wholeness,\\
Bending down on top of you, you now parted.\\
More letters from me now I read, more true,\\
Than even her gentle praise could cover,\\
Even of your loving lips.\\
To hear her speak, you must rely on her soft feature\\
To hear her speak well of loved ones.\\
\end{minipage}

\end{document}